%% file: arxiv_sub.tex
\documentclass{bmvc2k}
\usepackage{graphicx}

\usepackage{amsmath,amssymb} 
\usepackage{amsfonts} 
\usepackage{booktabs}
\usepackage{pifont}%
\usepackage{adjustbox}
\usepackage{multirow}
\usepackage{lscape}
\usepackage{longtable}

\usepackage{pifont}
\usepackage{color, colortbl}
\newcommand{\cmark}{\ding{51}}%
\newcommand{\xmark}{\ding{55}}%
\definecolor{darkmagenta}{RGB}{127,0,127}
\definecolor{LightCyan}{rgb}{0.88,1,1}
\definecolor{whitesmoke}{rgb}{0.96, 0.96, 0.96}
\newcommand{\tablelightgray}{\rowcolor[gray]{.95}}

\title{LOCL: Learning Object-Attribute Composition using Localization}

 \addauthor{Satish Kumar}{satishkumar@ucsb.edu}{1}
 \addauthor{ASM Iftekhar}{iftekhar@ucsb.edu}{1}
 \addauthor{Ekta Prashnani}{eprashnani@ucsb.edu}{1}
 \addauthor{B.S. Manjunath}{manj@ucsb.edu}{1}

\addinstitution{
ECE Department,\\
University of California\\
Santa Barbara}

\runninghead{Kumar et. al.}{LOCL}

\begin{document}

\maketitle
\input{Contents/1_abstract}
\vspace{-0.4cm}
\input{Contents/2_introduction}
\vspace{-0.4cm}
\input{Contents/3_relatedwork}
\vspace{-0.4cm}
\input{Contents/4_methodology}
\vspace{-0.4cm}
\input{Contents/5_experiments}
\vspace{-0.4cm}

\input{Contents/appendix_6_results}
\vspace{-0.45cm}
\input{Contents/7_conclusions}

\bibliography{egbib}

\input{Contents/appendix_arxiv}

\end{document}

%% file: Contents/1_abstract.tex
 \begin{abstract}
 This paper describes LOCL: Learning Object-Attribute (O-A) Composition using Localization – that generalizes composition zero shot learning to objects in cluttered/more realistic settings. The problem of unseen O-A associations has been well studied in the field, however, the performance of existing methods is limited in challenging scenes. In this context, our key contribution is a modular approach to localizing objects and attributes of interest in a weakly supervised context that generalizes robustly to unseen configurations.  Localization coupled with a composition classifier significantly outperforms state-of-the-art (SOTA) methods, with an improvement of about 12\% on currently available challenging datasets. Further, the modularity enables the use of localized feature extractor to be used with existing O-A compositional learning methods to improve their overall performance.

\end{abstract}

%% file: Contents/2_introduction.tex
\section{Introduction}
\vspace{-0.1cm}
\begin{figure}[t]
\begin{center}
\includegraphics[width=0.8\linewidth]{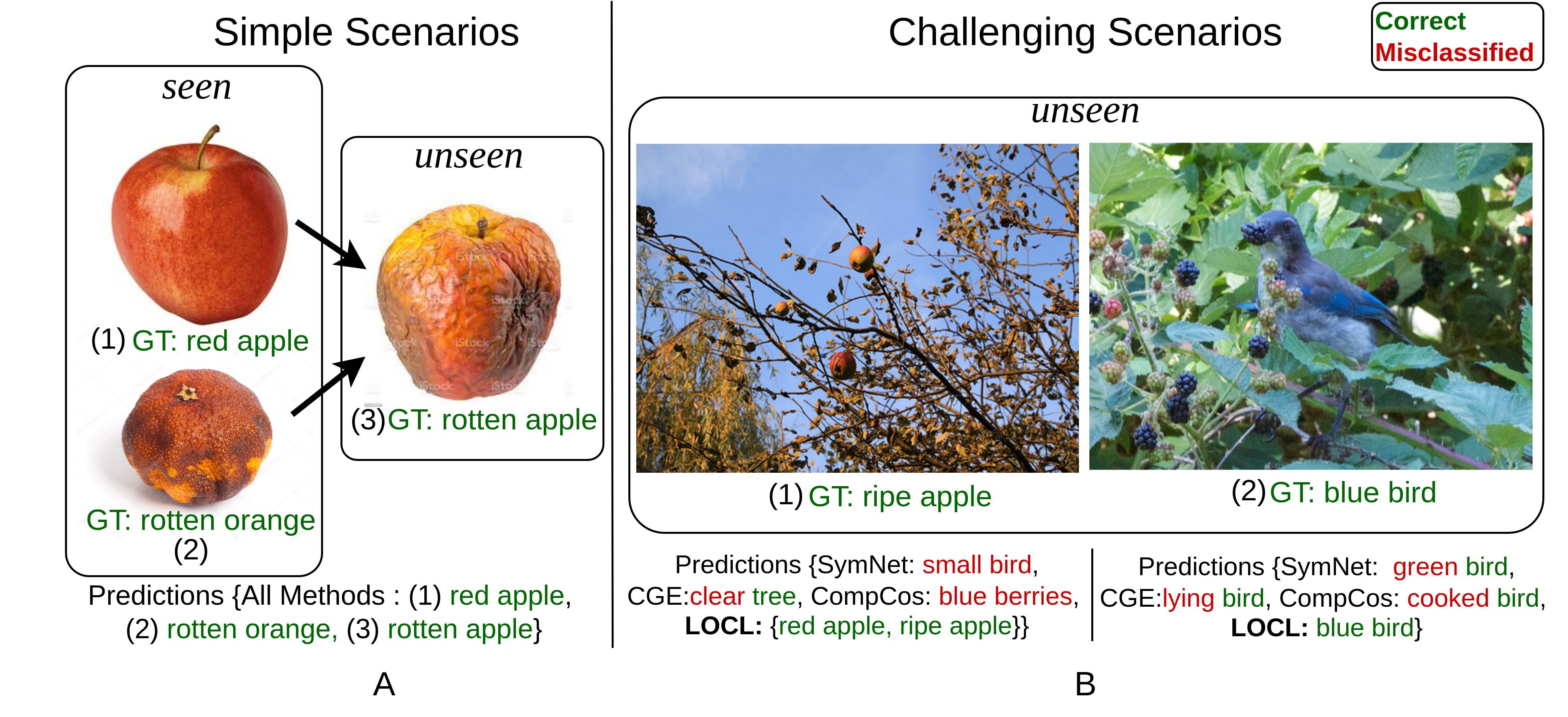}
\end{center}
 \vspace{-0.4cm}
\caption{The object of interest shown in images A.1, A.2, and A.3 presents simple scenarios where all SOTA
    (SymNet \cite{li2020symmetry}, CGE \cite{naeem2021learning}, CompCos \cite{mancini2021learning}) methods make correct O-A associations. However, for the same object (\textit{apple}) in a more cluttered scene in  image B.1, these methods fail. Even in cases where there is a dominant object of interest, such as a \textit{bird} in  (B.2), where there is significant background clutter, most of the SOTA methods have incorrect O-A associations.}
  \vspace{-0.2cm}
\label{fig:concept}
\end{figure} 

Human visual reasoning allows us to leverage prior visual experience to recognize previously unseen Object-Attribute (O-A) relationships. 
Predicting such complex relationships of novel O-A compositions -- referred to as Composition Zero Shot Learning (CZSL) \cite{mancini2021open,naeem2021learning,li2020symmetry, xu2021relation,ruis2021independent,misra2017red, Wei_2019_ICCV, purushwalkam2019task}--is an active area of research. There has been significant progress on CZSL methods in recent years, however, as our experiments demonstrate, their performance degrades in natural cluttered scenes, as illustrated in Fig.\ref{fig:concept}. 
The main reason in these cases is the interference from the other potential confusing elements. For example, in Fig. \ref{fig:concept}(B.1), the SOTA methods are not able to detect the object of interest given its size relative to image; and while the bird is the object of interest in Fig. \ref{fig:concept}(B.2), the surrounding context dominated by the green leaves results in an incorrect association of the color attribute to the object. 

The poor performance of the SOTA methods can be attributed to the dominant confounding elements thereby impeding the right O-A composition prediction. This in turn is due to the bias towards seen O-A composition during training time. Generalization to more realistic cases as seen in Fig. \ref{fig:concept}(B) is crucial for the widespread use of CZSL. 

Inspired by these limitations, we propose Learning Object-Attribute Composition using Localization (LOCL). Our model (LOCL) leverages spatially-localized learning, which is not present in the existing CZSL networks. It is reasonable to ask \textit{Why not first localize the objects and then associate the attributes?} In principle, this can be done, however, the SOTA methods for object detection and localization use extensive datasets for their training. Hence it will not be possible to meaningfully test the CZSL with pre-trained detectors. The images shown in Fig.~\ref{fig:concept} are from existing datasets for CZSL methods \cite{isola2015discovering, yu2017semantic, naeem2021learning}. \textit{We note that all the experiments reported in this paper use the datasets that are created for evaluating CZSL approaches.}

Existing SOTA object-attribute detection approaches do not take into account the possibility of scene attributes confounding with correct O-A composition prediction \cite{mancini2021open,naeem2021learning,li2020symmetry}. These methods are designed to work with wholistic image features \cite{xu2021relation,ruis2021independent,misra2017red, Wei_2019_ICCV, purushwalkam2019task}. Some recent work address this issue by partitioning the image into regions \cite{xu2015show, huynh2020compositional} or equal-size grid cells \cite{jaderberg2015spatial, zhao2019recognizing, huynh2020fine}, but they are not very effective in capturing distinctive object features.

Our approach towards better generalization of CZSL to more challenging images (Fig. \ref{fig:concept}.B) with background clutter is to leverage localized feature extraction in O-A composition. Specifically, we adopt a two step approach. First, a Localized Feature Extractor (LFE) associates an object with its attribute by reducing the interference arising from additional attribute-related visual cues occurring elsewhere in the image.  The CZSL benchmark datasets \textit{do not} contain any localization information. 
As noted before, off-the-shelf object detectors can be inadvertently exposed \cite{ren2015faster} to unseen OA compositions.
Therefore, we developed a weakly supervised method for localized feature extraction. Second, the composition classifier uses the localized distinctive visual features to predict an O-A pair.

The proposed LOCL outperforms competing CZSL methods on \textit{all} existing datasets -- including the more challenging CGQA -- providing a strong evidence in favor of its applicability to more realistic scenes.
Further, the performance of all existing methods improve when our localized feature extractor is included as a pre-processing module -- although LOCL still outperforms these methods.

%% file: Contents/3_relatedwork.tex
\section{Related Work}
\vspace{-0.2cm}
\begin{figure*}[t]
\begin{center}
\includegraphics[width=0.85\linewidth]{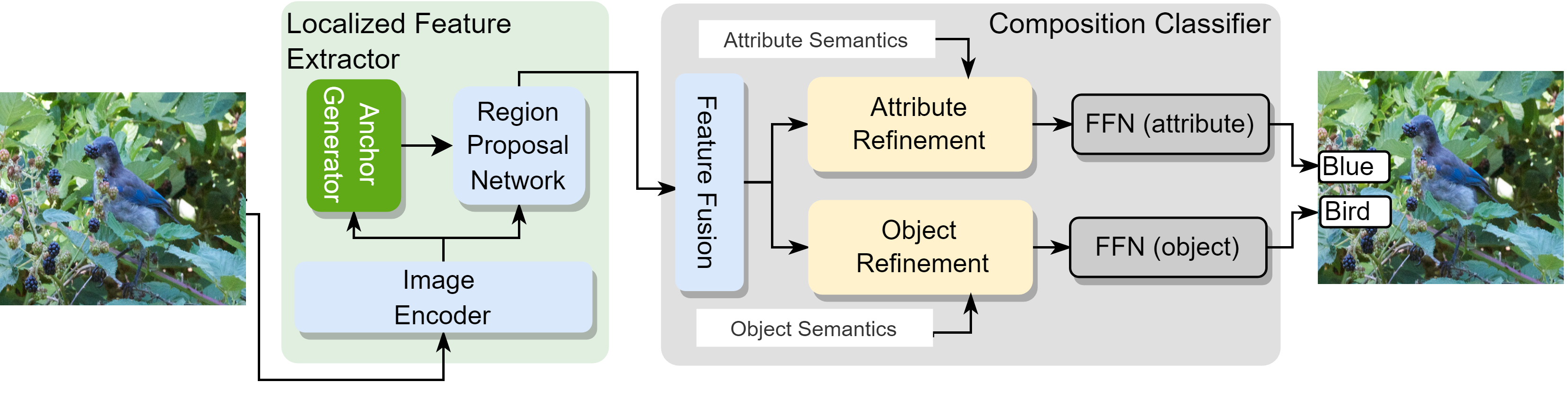}
\end{center}
    \vspace{-0.4cm}
   \caption{LOCL architecture. The Localized Feature Extractor (Section:~\ref{section:LFE}) generates proposals that are likely to contain objects. These proposals are refined with the object and attribute semantics using  Composition Classifier (Section:~\ref{section:CCL}). 
   }
\vspace{-0.2cm}
\label{fig:Architecture}
\end{figure*}

Existing work in object attribute (O-A) CZSL task typically assume that the images of the object of interest present in uncluttered context. This assumption is also true for the initial CZSL datasets~\cite{isola2015discovering,yu2017semantic}. As a result, most of the CZSL methods ~\cite{nan2019recognizing,mancini2021open,xu2021relation,ruis2021independent,misra2017red, Wei_2019_ICCV, purushwalkam2019task, Atzmon_casual,nagarajan2018attributes,naeem2021learning} perform quite well on uncluttered scenes. As noted before, some of the methods reduce the  interference from confounding elements by partitioning the image.
~\cite{zhao2019recognizing} is designed for datasets with a dominant object with a clear background.~\cite{huynh2020compositional} partitions the image to equal-sized grid cells and relies on aligning the attribute semantic and visual features. Further, the dataset used in~\cite{zhao2019recognizing,huynh2020compositional} consist of one object type,e.g., face or bird images.  

The O-A problem is considered as a matching problem in a latent space~\cite{misra2017red,Wei_2019_ICCV,nan2019recognizing, purushwalkam2019task}. For this matching task,~\cite{purushwalkam2019task} proposes a modular network with a dynamic gating whereas~\cite{misra2017red} defines objects and attributes using a support vector machine (SVM). On the other hand,~\cite{nagarajan2018attributes,li2020symmetry} consider attributes as functional operation over objects. More recently~\cite{Atzmon_casual,naeem2021learning,mancini2021open,xu2021relation,ruis2021independent} focus on the relationships among attributes and objects. ~\cite{Atzmon_casual} disentangle attributes and objects with metric based learning. ~\cite{naeem2021learning} learns attribute-object dependence in a semi-supervised graph structure where unseen combinations are considered connected during training.  This is extended in~\cite{mancini2021open, xu2021relation, vaswani2017attention} where all possible combination of objects and attributes are considered during inference. 

In general, the performance of current methods  drop significantly on images with background clutter. Taking inspiration from the generic pipeline of weakly supervised object detection (WSOD)~\cite{dietterich1997solving,gonthier2020multiple,uijlings2013selective,kumar2020deep,diba2017weakly,kantorov2016contextlocnet,mcever2022context}, LOCL consists of a region proposal network with pseudo-label generation module that leverages supervision from linguistic embeddings in a novel contrastive framework. This feature extraction can be utilized to improve the existing network's performance in images with background clutter as shown in Table~\ref{tab:ablation}. 

%% file: Contents/4_methodology.tex
\section{Approach}
\label{method} 
\vspace{-0.2cm}
The primary issue to be addressed is the scene complexity, that require that the methods are able to make the correct associations during the training phase and predict the unseen configuration during testing. The proposed method is intuitive and straightforward in creating a weakly supervised framework that is modular and generalizes well. The LOCL extracts localized features of object regions in the image, which allows it to learn useful OA relationships and also suppress spurious O-A relations from the background clutter. 

First, we pre-train a Localized Feature Extractor $LFE(.)$ (Sec.~\ref{section:LFE}) network to extract features from multiple regions of the image.
Second, the pre-trained $LFE(.)$ along with a Composition Classifier $CC(.)$ (Sec.~\ref{section:CCL}) network learns to detect the O-A composition. \textit{The key insight  is to leverage the features from regions containing the object of interest to learn accurate O-A associations}.  Fig.~\ref{fig:Architecture} summarizes the overall LOCL architecture.
\vspace{-0.2cm}
\paragraph{\textbf{Problem Setting:}} Let $\{I, T_o, T_a, (a, o)\}$ be the training dataset with $N$ samples, where $I$ is the input image. $T_o, T_a$ are the list of all object and attribute labels, respectively, and $(a$, $o)$ is the tuple of attribute-object pairs in the image. The O-A pairs labels used during training are categorised as seen pairs. The goal of CZSL trained model is to take in an input image $I$ and predict $(\hat{a}, \hat{o})$. The O-A pairs labels used during inference are novel and unseen. H\textit{ere the seen and unseen object-attribute pairs are mutually exclusive.}
\vspace{-0.2cm}
\paragraph{Proposed Network:} The proposed LOCL is as $(\hat{a}, \hat{o}) = CC\;(\;LFE\;(I),\; T_a, \;T_o)$,
where $LFE(.)$ and $CC(.)$ are trainable networks. LOCL is trained in two stages. In the first stage, given an image $I$, pre-training of $LFE(.)$ is done to generate multiple localized features. The details of the $LFE(.)$ module are discussed in Sec.~\ref{section:LFE}. The output of the trained network $LFE(.)$ is a list of $n$ features of object regions identified in the image.

In the second stage, out of these $n$ features, $r$ features ($r < n$) are input to the composition classifier ( Sec.~\ref{section:CCL}) to make the final prediction of attribute and object present in the image.

\begin{figure*}[t]
\begin{center}
\includegraphics[width=0.85\linewidth]{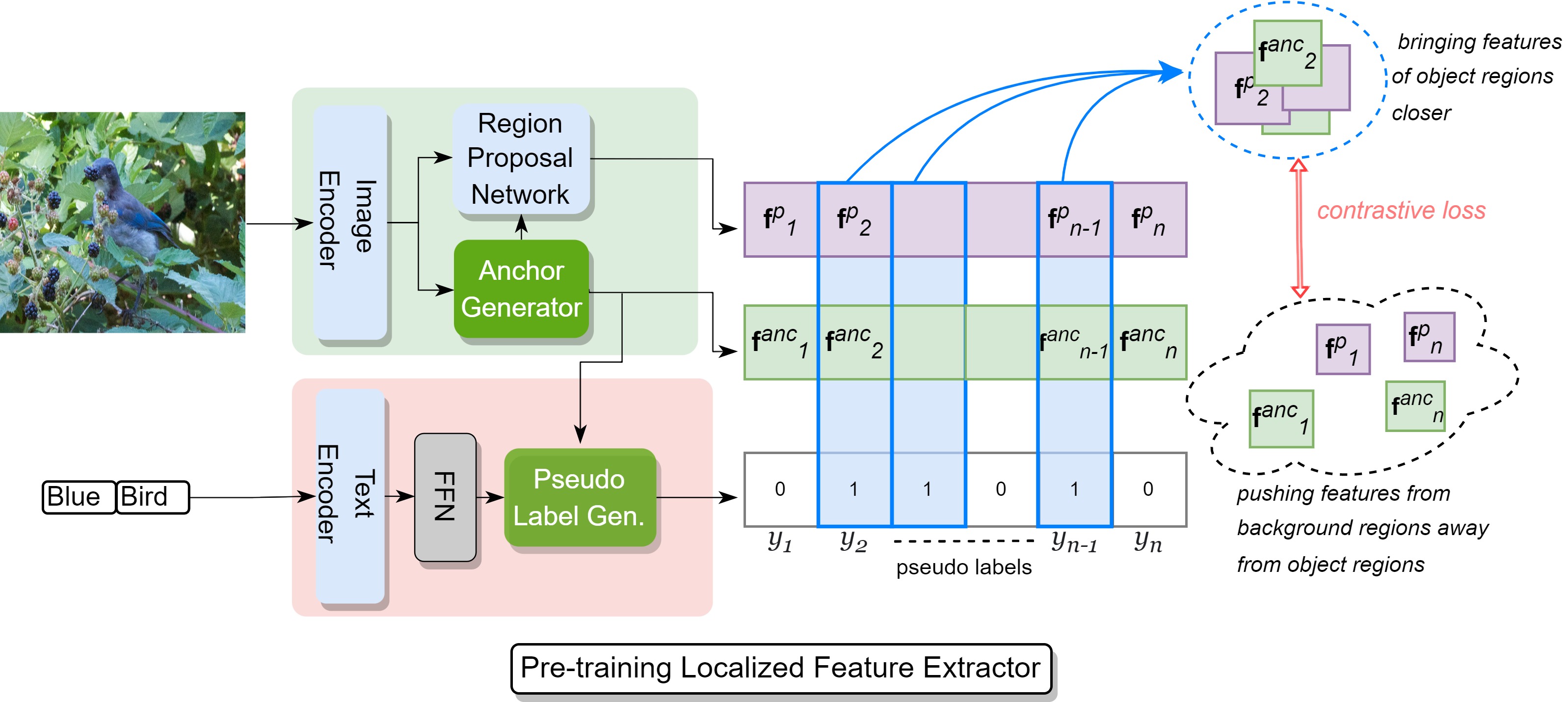}
\end{center}
 \vspace{-0.4cm}
   \caption{ Summary of pre-training the localized feature extractor.The image encoder and region proposal are jointly trained to generate features of object of interest. During training time, we use text embeddings to generate pseudo labels to train the image encoder and region proposal using contrastive learning. At the test time, the learned image encoder and region proposal network are used to generate features from object regions. 
   }
\label{fig:ALM}
\vspace{-0.3cm}
\end{figure*}

\vspace{-0.3cm}
\subsection{Pre-training Localized Feature Extractor($LFE(.)$)}\label{section:LFE}
\vspace{-0.2cm}
Our Localized Feature Extractor network $LFE(.)$ is a combination of an image encoder (ResNet-50~\cite{he2016deep}), text encoder, Region Proposal Network (RPN), and a pseudo label generator, as shown in Fig.~\ref{fig:ALM}. 
The RPN is inspired from F-RCNN~\cite{ren2015faster}. It is trained from scratch using a contrastive learning framework.
It generates proposals features for regions in the image that has high likelihood of object presence.
The pseudo label generator creates labels to supervise the visual space that have high semantic similarity with ground truth O-A pair.

Given the input image $I$, the image encoder generates feature map $\textbf{F} \in \; \mathbb{R}^{H'\times W'\times C}$ where $H', W'$ and $C$ are the height, width and channel dimensions. Then $n$ valid anchors (based on input image size, in our case for an image of $256 \times 256$, $n=576$,) are generated on the input image~\cite{ren2015faster}. Anchors are a set of rectangular boxes with different aspect ratio and scale generated at \textbf{each pixel of the input image}~\cite{ren2015faster}. Corresponding to each anchor, a list of features is pooled from $\textbf{F}$. The pooled anchor features are $[\textbf{f}^{\;anc}_1, \textbf{f}^{\;anc}_2, ..., \textbf{f}^{\;anc}_n]  \in \; \mathbb{R}^{C}$ shown as output of ``Anchor Generator" in Fig.~\ref{fig:ALM}

The text encoder generates semantic pair embedding from the input text label $(a:Blue$, $o:Bird)$ pair. With the help of these semantic pair embedding, we generate pseudo ground truth labels to train $LFE(.)$ network with weak supervision~\cite{tian2019contrastive}. In the following, we refer to ``pseudo ground truth labels" as ``pseudo labels" for simplicity.

\vspace{-0.35cm}
\paragraph{Pseudo Label Generator:} The ground truth O-A semantic pair embeddings generated from text encoder are projected through fully connected layers ($FFN$) into a common subspace as visual embeddings. The output of $FFN$ is denoted by $\textbf{f}^{\;text}_{ao} \in \; \mathbb{R}^{C}$, where ``$ao$" index is the ground truth (in our case one per image). Here the length $C$ of semantic embedding equal to channel dimension $C$ of visual feature vector \textbf{F}. Now to generate pseudo labels, a cosine similarity score is computed between each $\textbf{f}^{\;anc}_k$ and $\textbf{f}^{\;text}_{ao}$.
\begin{equation}
    \begin{split}
        \phi_{k} = \frac{\textbf{f}^ {\;text}_{ao}\cdot\;\textbf{f}^{\;anc}_k}{||\textbf{f}^{\;text}_{ao}||\;||\textbf{f}^{\;anc}_k||}\; \; \forall \;\; \textbf{f}^{\;anc}_k, \;where\;\; (\phi = [\phi_1, \phi_2, .., \phi_k, .., \phi_n])
    \end {split}
    \label{eq:pseudo_label1}
\end{equation}
\vspace{-0.2cm}
\begin{equation}
  \textbf{y} =
    \begin{cases}
      1 & argsort(\phi)[0:l]\\
      0 & for\; all\; other\;indexes\\
    \end{cases}
    \label{eq:pseudo_label}
\end{equation}
where $\mathbf{y} = [y_1, y_2, .., y_k, .., y_n]$, $l$ top anchors are selected out of $n$ based on cosine similarity score $\phi$. 
They are assigned with label 1 in $\mathbf{y}$ and rest are assigned 0 as shown above with Eq.~\ref{eq:pseudo_label}. Here each $y_k$ represents the presence/absence of object of interest regions in the image.
Intuition here is that $\textbf{f}^{\;anc}_k$'s which contains the object will lie closer to $\textbf{f}^{\;text}_{ao}$ in feature space.

\vspace{-0.2cm}
\paragraph{Region proposal Network (RPN):} The RPN branch shown in Fig.~\ref{fig:ALM} is inspired from FasterRCNN~\cite{ren2015faster}. 
The RPN generated proposals are used to pool a list of features from the feature map $\textbf{F}$. The pooled features are  $[\textbf{f}^{\;p}_1, \textbf{f}^{\;p}_2, ..., \textbf{f}^{\;p}_n]  \in \; \mathbb{R}^{C}$. Now the anchor features $[\textbf{f}^{\;anc}_1, \textbf{f}^{\;anc}_2, ..., \textbf{f}^{\;anc}_n]$, proposal features $[\textbf{f}^{\;p}_1, \textbf{f}^{\;p}_2, ..., \textbf{f}^{\;p}_n]$ and pseudo label $y=[y_1, y_2, .., y_n]$ are used to train the function $LFE(.)$ using contrastive learning as explained in next section.

\vspace{-0.2cm}
\paragraph{Contrastive Pre-training:}\label{section:contrastive pre-train}Recall that the current benchmark CZSL datasets~\cite{isola2015discovering,mancini2021learning,yu2017semantic} do not have ground truth bounding boxes for objects of interest. For this reason, we use both the anchor features and proposal features to localize the objects of interest. This is different from regular object detection networks~\cite{ren2015faster}. 
The pseudo label $\mathbf{y}$ informs the network which anchor features are likely to represent object(s) of interest. Using contrastive learning, we train the regional proposal network branch to localize the object with weak supervision.
The goal of contrastive learning is to maximize the similarity between similar feature vectors and minimize the similarity between the dissimilar feature vectors. 
Here, the objective is to maximize the cosine similarity between $< \textbf{f}^{\;anc}_k, \textbf{f}^{\;p}_k>$ features where there is a possibility of object being present and minimize in all other cases. The contrastive objective function is:

\begin{equation}
    \mathcal{L}_{CON}= \sum_{\substack{k=1}}^{n}(1-y_{k})*d_{k}^2
    + y_{k}*{\max (0,\: 1 - d_{k}^2)},
    \label{eq:CL}
\end{equation}
where $*$ is element wise multiplication, $y_{k}$ tells us which features have the possibility of having an object (Eq.~\ref{eq:pseudo_label}), $d_{k}$ is the cosine distance between $< \textbf{f}^{\;anc}_k, \textbf{f}^{\;p}_k>$.
\begin{equation}
    d_{k} = \frac{\textbf{f}^{\;p}_k \cdot \textbf{f}^{\;anc}_k}{||\textbf{f}^{\;p}_k||\;\; ||\textbf{f}^{\;anc}_k||} \; \forall\; k=[1, 2, .. n]     
\end{equation}
Along with contrastive loss, we optimize binary cross entropy over the objectness score predicted by region proposal network. The overall loss function is:
\begin{equation}
    \mathcal{L}_{total} = \alpha * \mathcal{L}_{CON} + \beta * \mathcal{L}_{BCE}(o, \phi),
    \label{eq:loss_cl}
\end{equation}
where, $\alpha$ and $\beta$ are empirically-determined scaling parameters, $o$ is the objectness score from RPN and $\phi$ is the cosine distance from Eq.~\ref{eq:pseudo_label1}. Once the $LFE(.)$ network is trained, the output of trained model are the proposal feature vectors $[\hat{\textbf{f}^{\;p}_1}, \hat{\textbf{f}^{\;p}_2}, ..., \hat{\textbf{f}^{\;p}_n}]$ along with objectness score $\hat{o} = [\hat{o_1}, \hat{o_2}, .., \hat{o_n}]$. This learnt parameter objectness score ensures selection of features with object information, thereby minimizing the interference from potential confusing elements as shown in Fig.~\ref{fig:concept}

\vspace{-0.2cm}
\subsection{Composition Classifier $CC(.)$}\label{section:CCL}
\begin{figure}
    \centering
    \includegraphics[width=0.75\linewidth]{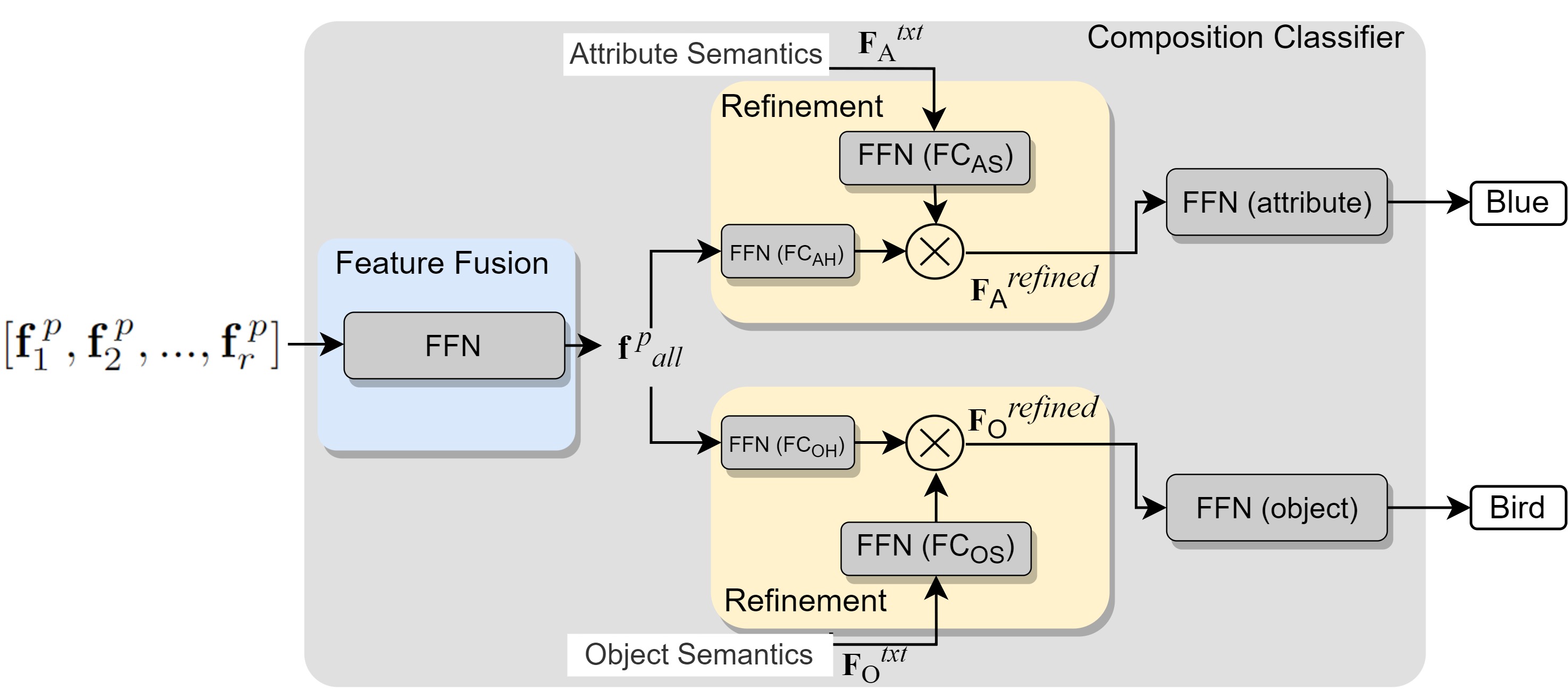}
    \vspace{-0.2cm}
    \caption{Composition Classifier $CC(.)$ architecture. The proposal features $[\hat{\textbf{f}^{\;p}_1}, \hat{\textbf{f}^{\;p}_2}, ..., \hat{\textbf{f}^{\;p}_r}]$ are the outputs from $LFE(.)$ which are combined into a single representation $\textbf{f}^{\;p}_{all}$. The attribute and object semantics are the semantic encoding of all attributes and objects under consideration. Two branches predict attribute and object from semantically refined $\textbf{f}^{\;p}_{all}$.}
    \vspace{-0.4cm}
    \label{fig:CC}
\end{figure}

\vspace{-0.15cm}
 The ability to learn individual representation of O-A in visual domain is crucial for transferring knowledge from seen to unseen O-A associations. Existing SOTA works~\cite{naeem2021learning,mancini2021learning,mancini2021open,ruis2021independent,xu2021zero} use homogeneous features from whole image as without localizing the object, they ignore the discriminative visual features of object and its attributes. Our Composition Classifier network $CC(.)$ leverages the distinctive features extracted by $LFE(.)$ to predict the object and its corresponding attribute as shown in Fig.~\ref{fig:CC}.
 It is challenging to associate right attribute with the object by using homogeneous features, as there can be interference from prominent confounding elements like examples shown in Fig\ref{fig:concept}B. Similarly in Fig.~\ref{fig:Qual-Results1} (row-2, column-1 image), the attribute ``$green\; color$" is so prominent, and using homogeneous features can lead to wrong O-A prediction.

The input to $CC(.)$  is a set of top $r$ ($r < n$) proposal feature vectors from pre-trained $LFE(.)$ $[\textbf{f}^{\;p}_1, \textbf{f}^{\;p}_2, ..., \textbf{f}^{\;p}_r] \; \in \mathbb{R}^{r \times C}$ sorted in descending order based on objectness score $o = [o_1, o_2, .., o_r]$. The proposal feature vectors are fused into a single visual feature $\textbf{f}^{\;p}_{all} \in \mathbb{R}^{1 \times C} $ using weighted average with learnable parameters~\cite{kumar2021stressnet} as shown in Fig.~\ref{fig:CC}.
Input to the learnable parameters has $r \times C$ dimension. $r$ is the number of proposals and $C$ is the number of channels. We swap the dimensions at the input to fuse different proposals together. The output of it is a single feature vector of length $C$.
The fusion operation is a learnable weighted average operation, that learns to create joint representation from features of object regions. 
Next, $\textbf{f}^{\;p}_{all}$ is projected to two different representation via two feed forwards networks $FC_{AH}$ and $FC_{OH}$ layers as shown in Fig.~\ref{fig:CC}. 
Similarly, the semantic embeddings of all attributes $\textbf{F}^{\;text}_A = [\textbf{f}^{\;text}_{a1}, \textbf{f}^{\;text}_{a2}.., \textbf{f}^{\;text}_{ai}]$ and all objects $\textbf{F}^{text}_O = [\textbf{f}^{\;text}_{o1}, \textbf{f}^{\;text}_{o2}.., \textbf{f}^{\;text}_{oj}]$ (generated from $T_a$ and $T_o$) are projected via feed forward network $FC_{AS}$ and $FC_{OS}$.
Here $T_a$ and $T_o$ are the list of all attributes and objects in the dataset respectively, and $i = len(T_a),\; j = len(T_o)$. $len(.)$ is length.
Inspired by ~\cite{vaswani2017attention,ulutan2020vsgnet,li2020symmetry,iftekhar2021gtnet,iftekhar2022look}, the visual feature projections are refined as shown below. 
The particular choice of refinement by one-to-one multiplication is an empirical choice. We explore different refinement techniques in Table 3 of supplementary materials.
This all is done by $Refinement$ block as shown in Fig.~\ref{fig:CC}.
\begin{equation}
    \textbf{F}^{refined}_O = FC_{OH}(\textbf{f}^{\;all}_p) \circ FC_{OS}(\textbf{F}^{txt}_O)
    \;\; ;\;\; 
    \textbf{F}^{refined}_A = FC_{AH}(\textbf{f}^{\;all}_p) \circ FC_{AS}(\textbf{F}^{txt}_A)
    \label{eq:refine_qa}
\end{equation}
where, ``$\circ$" represents element-wise multiplication and the feed forward network are two fully connected layers with ReLu activation.
This refinement aggregates semantic information with the visual features. These refined features are passed through another feed forward network (Fig.~\ref{fig:CC}) and softmax layers to make the final decision on the object $\hat{o}$ and attribute $\hat{a}$ present in the image $I$. To train the composition classifier function $CC(.)$, we optimize binary cross entropy function over the O-A prediction.


%% file: Contents/5_experiments.tex
\section{Experiments}\label{section:implementation}
\vspace{-0.15cm}
Table~\ref{tab:datasets} summarizes the datasets used. In the MIT-states dataset the images are of natural objects collected using an older search engine with limited human annotation causing a significant label noise~\cite{Atzmon_casual}. For UT-Zappos~\cite{yu2017semantic} the simplicity of the images (one object with white background) makes it unsuitable to work in natural surroundings. We performed experiments on very recently released Compositional GQA (CGQA) dataset~\cite{mancini2021learning,hudson2019gqa}. This dataset is proposed to evaluate existing CZSL models in a more realistic challenging scenarios with background clutter. More details about the datasets can be found in supplementary materials section 4. 

Both the localized feature extractor $LFE(.)$ and Composition Classifier $CC(.)$ are trained on all the datasets. To pre-train $LFE(.)$, an efficient contrastive pre-training framework is used with a margin distance of 1. Then this pre-trained network is used with $CC(.)$ to do an end-to-end training of LOCL. We have provided the complete implementation details in the supplementary materials. However, during inference time, $LFE(.)$ do not have any text embedding branch, hence it generates proposals for every potential object as shown in Fig.~\ref{fig:samplebb2}. 

Following current methods~\cite{purushwalkam2019task,li2020symmetry,naeem2021learning}, we evaluate our network's performance in Generalized Compositional Zero Shot Learning Protocol (GCZSL). Under this protocol, we draw seen class accuracy vs unseen class accuracy curve at different operating points of the network. These operating points are selected from a single calibration scaler that is added to our network's predictions of the unseen classes~\cite{purushwalkam2019task, li2020symmetry, naeem2021learning}.   
We report area under ``seen class accuracy vs unseen class accuracy curve" (AUC) as our performance metrics. Additionally, we report our network's performance on Top-1 accuracy in seen and unseen classes.

\begin{table}[]
\centering
\adjustbox{width=0.85\columnwidth}{
\begin{tabular}{c|ccc|c|c|c}
\multicolumn{1}{l|}{}                          & \multicolumn{3}{c|}{\# Images} & \# Objects & \#Attributes & \# OA Pairs \\
\multicolumn{1}{l|}{}                          & Train     & Val     & Test     &            &              &             \\ \hline
\tablelightgray MIT-States~\cite{misra2017red} & 30k       & 10k     & 13k      & 245        & 115          & 1962        \\
UT-Zappos~\cite{yu2017semantic}                & 23k       & 3k      & 3k       & 12         & 16           & 116         \\
\tablelightgray CGQA~\cite{naeem2021learning}  & 26k       & 7k      & 5k       & 870        & 453          & 9378       
\end{tabular}
}
\vspace{0.2cm}
\label{tab:datasets}

\caption{Comparison of different CZSL datasets~\cite{misra2017red, yu2017semantic, naeem2021learning}}
\label{tab:datasets}
\vspace{-0.1cm}
\end{table}


%% file: Contents/appendix_6_results.tex
\subsection{Results} \label{section:results}
LOCL outperforms current methods in the test set of all benchmark datasets in almost all categories as shown in Table~\ref{tab:main_result}. We evaluate LOCL's performance in terms of AUC under Generalized Compositional Zero Shot Learning (GCZSL) protocol. 
We also report Top-1 accuracy in seen and unseen classes and accuracy in detecting objects and attributes. 
In MIT-States~\cite{isola2015discovering} LOCL outperforms SOTA method by $8\%$ on unseen class accuracy and $1.7\%$ AUC. 
In UT-Zappos~\cite{yu2017semantic} LOCL's unseen class accuracy is $5.2\%$ better than the SOTA method. Moreover, it almost \textit{doubles} the unseen class accuracy while achieving $1.1\%$ improvement in terms of AUC for the more challenging CGQA~\cite{naeem2021learning} dataset. 

Current CZSL methods use homogeneous features from the backbone instead of using distinctive visual features of objects and attributes. While such techniques may work on simpler datasets like UT-Zappos~\cite{yu2017semantic}, as evidenced by the high performance, the more realistic datasets such as CGQA pose challenges. 
Table~\ref{tab:main_result} shows, LOCL achieves the best results on the challenging CGQA dataset. 
However, bias towards seen O-A compositions is a common issue~\cite{purushwalkam2019task} in current CZSL methods. Recent approaches~\cite{xu2021relation, naeem2021learning} have utilized a graph structure with message passing/blocking~\cite{xu2021relation} or prior possible O-A knowledge~\cite{naeem2021learning} to reduce this bias. 
However they tend to be biased towards seen O-A pairs at inference as pointed out by the authors of ~\cite{xu2021relation}.
In contrast, LOCL learns distinct object and attribute representations in the two separate branches of the $CC(.)$ and achieves high unseen class accuracy and AUC.
In UT-Zappos, high AUC of~\cite{xu2021relation} stems from high seen class accuracy with inferior unseen class performance.~\cite{xu2021relation} do not evaluate their model CGQA dataset~\cite{naeem2021learning}.

\begin{table*}[t]
\centering
\adjustbox{width=\textwidth}{
\begin{tabular}{l|ccc|ccc|ccc}
\multicolumn{1}{c|}{\multirow{2}{*}{Methods}}        & \multicolumn{3}{c|}{MIT-States~\cite{isola2015discovering}} & \multicolumn{3}{c|}{UT-Zappos~\cite{yu2017semantic}} & \multicolumn{3}{c}{CGQA~\cite{naeem2021learning}} \\ 
\multicolumn{1}{c|}{}                                & Seen               & Unseen             & AUC               & Seen             & Unseen          & AUC             & Seen            & Unseen          & AUC           \\ \hline
\tablelightgray Attop~\cite{nagarajan2018attributes} & 14.3               & 17.4               & 1.6               & 59.8             & 54.2            & 25.9            & 11.8            & 3.9             & 0.3           \\
LabelEmbed~\cite{misra2017red}                       & 15                 & 20.1               & 2.0               & 53.0             & 61.9            & 25.7            & 16.1            & 5               & 0.6           \\
\tablelightgray TMN~\cite{purushwalkam2019task}      & 20.2               & 20.1               & 2.9               & 58.7             & 60.0            & 29.3            & 21.6            & 6.3             & 1.1           \\
SymNet~\cite{li2020symmetry}                         & 24.2               & 25.2               & 3.0               & 49.8             & 57.4            & 23.4            & 25.2            & 9.2             & 1.8           \\
\tablelightgray CompCos~\cite{mancini2021open}       & 25.3               & 24.6               & 4.5               & 59.8             & 62.5            & 28.1            & 28.1            & 11.2            & 2.6           \\
ProtoProp~\cite{ruis2021independent}                 & -                  & -                  & -                 & 62.1             & 65.5            & 34.7            & 26.4            & 18.1            & 3.7           \\
\tablelightgray BMP-Net~\cite{xu2021relation}        & \textbf{38.6}      & 21.7               & 6.0               & \textbf{87.3}    & 64.5            & \textbf{49.7}   & -               & -               & -             \\
CGE~\cite{naeem2021learning}                         & 32.8               & 28.0               & 6.5               & 64.5             & 71.5            & 33.5            & \textbf{31.4}   & 14              & 3.6           \\
\tablelightgray LOCL (Ours)                          & 35.3               & \textbf{36.0}      & \textbf{7.7}      & 68.0             & \textbf{76.7}   & 37.9            & 29.6            & \textbf{26.4}   & \textbf{4.2}  \\ \hline
\end{tabular}

}
\vspace{0.1cm}
 \caption{Performance comparisons on MIT-States~\cite{isola2015discovering}, UT-Zappos~\cite{yu2017semantic}, CGQA~\cite{naeem2021learning} Datasets. `-' means unreported performance in a particular category. In all three datasets, LOCL significantly outperform current methods. Specially, for the more challenging (significant background clutter) CGQA dataset, the effectiveness of LFE is clearly demonstrated by its performance on the unseen O-A associations.}
\label{tab:main_result}
\vspace{-0.3cm}
\end{table*}

\begin{table}[t]
\centering
\adjustbox{width=0.85\linewidth}{
\begin{tabular}{ccc|ccc|ccc}

\hline
\multirow{2}{*}{Methods} & \multirow{2}{*}{\begin{tabular}[c]{@{}c@{}}Our\\ BB\end{tabular}} & \multirow{2}{*}{LFE} & \multicolumn{3}{c|}{CGQA~\cite{naeem2021learning}}                                                        & \multicolumn{3}{c}{MIT-States~\cite{isola2015discovering}}                                                  \\ 
                         &                                                                         &                      & \multicolumn{1}{l}{Seen} & \multicolumn{1}{l}{Unseen} & \multicolumn{1}{l|}{AUC} & \multicolumn{1}{l}{Seen} & \multicolumn{1}{l}{Unseen} & \multicolumn{1}{l}{AUC} \\ \hline
\tablelightgray
\multirow{3}{*}{SymNet~\cite{li2020symmetry}}  & \xmark                                                                  & \xmark               & 25.2                     & 9.2                       & 1.8                      & 24.2                     & 25.2                       & 3.0                        \\
                         & \cmark                                                                  & \xmark               & 25.3                     & 9.3                       & 1.8                      & 26.6                     & 26.1                       & 3.5                      \\
                         \tablelightgray& \cmark                                                                  & \cmark               & 27.7                     & 13.5                       & 2.0                      & 28.7                     & 27.7                       & 3.8                      \\ \hline
\tablelightgray
\multirow{3}{*}{CompCos~\cite{mancini2021learning}} & \xmark                                                                  & \xmark               & 28.1                     & 11.2                       & 2.6                      & 25.3                     & 24.6                       & 4.5                      \\
                         & \cmark                                                                  & \xmark               & 28.4                       & 13.5                       & 2.8                      & 25.6                     & 24.8                       & 4.5                      \\
                         \tablelightgray& \cmark                                                                  & \cmark               & 28.9                     & 16.7                      & 2.9                      & 27.9                     & 26.7                       & 5.1                      \\ \hline
\tablelightgray
\multirow{3}{*}{CGE~\cite{naeem2021learning}}     & \xmark                                                                  & \xmark               & 31.4                     & 14.0                       & 3.6                      & 32.8                     & 28                         & 6.5                      \\
                         & \cmark                                                                  & \xmark               & 31.4                     & 19.3                      & 3.8                      & 33.3                     & 28                         & 6.5                      \\
                         \tablelightgray& \cmark                                                                  & \cmark               & \textbf{31.9}                     & 26.1                       & 4.1                     & \textbf{36.3}                     & 29.8                       & 6.6                      \\ \hline
LOCL                    & \cmark                                                                  & \cmark               & 29.6            & \textbf{26.4}              & \textbf{4.2}            & 35.3            & \textbf{36.0}              & \textbf{7.7}             \\ \hline
\end{tabular}
}
\vspace{0.2cm}
\caption{Performance of SOTA methods with our backbone (BB) and LFE. LFE significantly boosts all SOTA network performances specially in the CGQA dataset. Row 2 if each methods shows performance of all SOTA models with a common and better backbone.}
\label{tab:ablation}
\vspace{-0.3cm}

\end{table}

\vspace{-0.1cm}
\subsection{Ablation Study}
\vspace{-0.1cm}
\begin{figure*}[h]
\begin{center}
\includegraphics[width=0.8\linewidth]{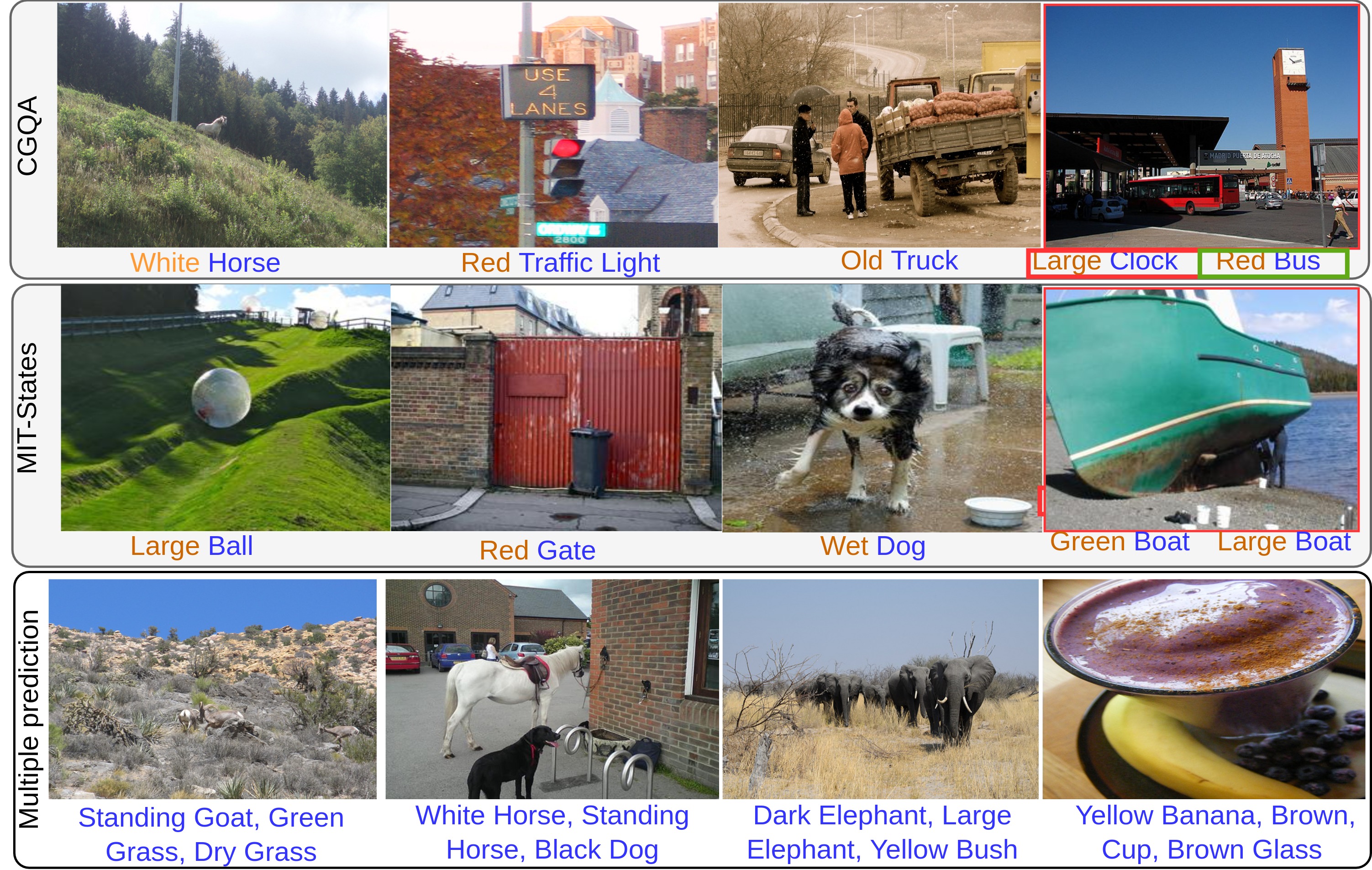}
\end{center}
\vspace{-0.4cm}
   \caption{Qualitative results of LOCL. Row-1 \& 2 (col-1,2,3)show correct predictions. Col-4 shows missed predictions, ground truth labels are marked with green box and our predictions in red box. The datasets has one O-A pair per image. Though our predictions are visually correct, do not match the ground-truth. This puts an artificial limit on the evaluation metric. Row-3 shows multiple O-A detections}
\vspace{-0.2cm}
\label{fig:Qual-Results1}
\end{figure*}


\noindent{\textbf{Backbone and Localized Feature Extractor:}} Our Localized Feature Extractor $LFE(.)$ is modular and can easily be adapted to other methods. In Table~\ref{tab:ablation}, we show different SOTA methods' improved performances with our feature extraction. 
For the sake of fair comparison with SOTA methods: SymNet~\cite{li2020symmetry}, ComCos~\cite{mancini2021learning} and CGE~\cite{naeem2021learning}, we replace their backbones with our ResNet-50 pre-trained on a larger dataset~\cite{radford2021learning}. 
As expected, both our backbone and $LFE(.)$ improve the existing networks' performances. This shows that the performance improvement is because of localized features generated from $LFE(.)$.
All existing methods use ResNet-18 as the backbone following the seminal work of~\cite{misra2017red}. 
We recommend that CZSL networks should utilize stronger backbones for challenging datasets like CGQA~\cite{naeem2021learning}. 
However, our improved performance is not just coming from a stronger backbone. 
With $LFE(.)$, the performance boost is more significant (specially in terms of unseen class accuracy) than the performance boost with our backbone for the CGQA dataset. 
In particular, $LFE(.)$ increases the unseen class accuracy of CGE~\cite{naeem2021learning} in CGQA by $\emph{86\%}$, and other methods also get great improvement with $LFE(.)$. 
The performance improvement in the MIT-States dataset is less due to the noisy annotations~\cite{Atzmon_casual} of this dataset.  
In summary, $LFE(.)$ improves three different architectures thus proving the effectiveness of localized feature extraction. 

We additionally ablate LOCL's performance for different \textit{number of proposals}, \textit{number of pseudo labels}, \textit{refinement techniques}, \textit{margin distance for contrastive loss}, and \textit{scaling parameters $\alpha$ and $\beta$}. All these experiments along with LOCL's performance on detecting individual objects and attributes are reported in the appendix section. 



\begin{figure}[t]
\begin{center}
\includegraphics[width=\linewidth]{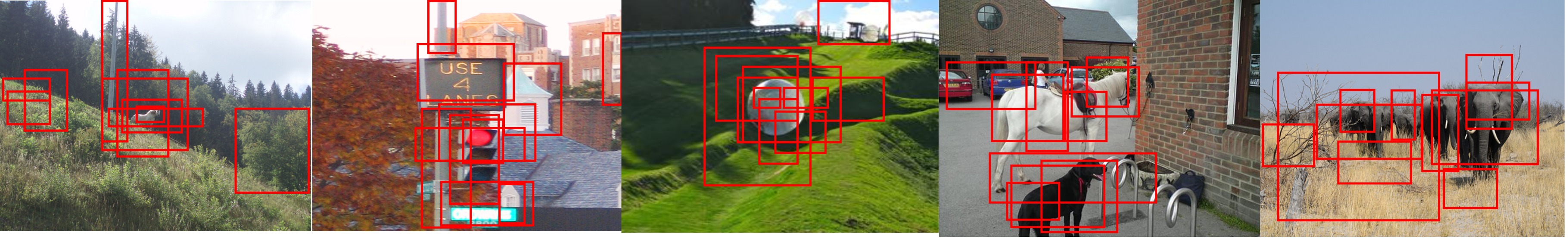}
\end{center}
  \vspace{-0.25in}
   \caption{Proposals selected based on objectness score. We can see that the proposals are generated on the object of interest. Though LOCL is not designed for multi O-A, but in case of multiple objects, the proposals are distributed over multiple objects.}
   \vspace{-0.25cm}
\label{fig:samplebb2}
\end{figure}

\vspace{-0.3cm}
\paragraph{Qualitative Results:}
We show results for unseen novel compositions with top-1 prediction in Fig.~\ref{fig:Qual-Results1}. They represent the scenes with clutter or confounding elements.
For example, column-1 shows where the confounding element $color$ attribute $<green>$ causes wrong O-A association for most of the SOTA methods while LOCL makes the right association. Similar in the first row.
The last column shows where our network predictions do not match with the ground truth labels. 
In the top image our network focuses more on prominent object $clock$, while the image is labeled for $<red, bus>$.
For row-2 image, it contains attribute categories like $size$, $color$, $texture$ but the label only has attribute $size$. We should, however, note that the ground truth labels in these datasets contain only one O-A pair. 
This puts an artificial limitation on the evaluation metric even when the predictions are \textit{perceivably correct} but does not match labels.
We also show $LFE(.)$ proposals quality in eliminating background in Fig.~\ref{fig:samplebb2}

\vspace{-0.2cm}

\paragraph{Multi O-A prediction:}
We extended LOCL to unconstrained setting than existing evaluation methods allows, i.e. detecting multiple O-A pairs. LOCL has flexibility of establishing the right O-A association for multiple objects in the scene. We are showing the top-3 O-A pairs prediction in Fig.~\ref{fig:Qual-Results1} last row. As can been seen the prediction are perceivably correct but unavailability of ground-truth annotation in existing datasets puts a limit on quantitative evaluation. Also, there can be multiple correct attribute related to one object as shown in the two central images in Fig.~\ref{fig:Qual-Results1} last row. $<White>, <Standing>$ both are correct attribute for object $<Horse>$, similarly $<Dark>, <Large>$ are correct attributes for object $<Elephant>$. Multiple OA pairs detection is an interesting direction for future research.

%% file: Contents/7_conclusions.tex
\section{Conclusion}
\vspace{-0.2cm}
We present a novel two-step approach, LOCL, for recognizing O-A pairs. Our approach includes a robust local feature extractor followed by a composition classifier.
LOCL is evaluated on benchmark datasets. Additionally, our experiments show that the local feature extractor improves the performance of current SOTA CZSL models by a significant margin of 12\%. The code and trained model will be made available on Github at the time of publication.  

\section{Acknowledgement}
\vspace{-0.2cm}
The authors would like to thank Dr. Suya You, Army Research Laboratory (ARL), for the many discussions that contributed to the methods development. We also thank Mr. Pushkar Shukla from Toyota Technological Institute, Chicago and Mr. Rahul Vishwakarma from Hitachi Labs America for their suggestions and critical reviews of the paper. This research was in part supported by NSF SI2-SSI award \#1664172 and  by US Army Research Laboratory (ARL) under agreement \# W911NF2020157. The U.S. Government is authorized to reproduce and distribute reprints for Governmental purposes notwithstanding any copyright notation thereon. The views and conclusions contained herein are those of the authors and should not be interpreted as necessarily representing the official policies or endorsements, either expressed or implied, of US Army Research Laboratory (ARL) or the U.S. Government

%% file: Contents/appendix_arxiv.tex
\section{Appendix}

\subsection{Implementation Details}

Both the localized feature extractor $LFE(.)$ and Composition Classifier $CC(.)$ are trained on all the datasets. To train $LFE(.)$, an efficient contrastive pre-training framework is used with a margin distance of 1. As backbone image encoder, we use ResNet-50~\cite{he2016deep} pre-trained on~\cite{radford2021learning}. For text encoding we utilize text encoder similar to~\cite{radford2021learning}. The Anchor Generator generates 576 valid anchor boxes. Corresponding to each of these anchor boxes, features $[\textbf{f}^{\;anc}_1, \textbf{f}^{\;anc}_2, ..., \textbf{f}^{\;anc}_{576}]$ are pooled from $\textbf{F}$. To generate $\phi$ according to Eq.~\ref{eq:pseudo_label2}, each $\textbf{f}^{\;anc}_j$ is matched with semantic word embedding vector, and top 20 scores are labeled as 1 and rest as 0 as shown in Eq.~\ref{eq:pseudo_label} to create the pseudo label $y$. This is an empirically selected value, it covers almost all the object regions in the image. The \textbf{Region Proposal Network} generates proposal boxes and an objectness score corresponding to each proposal box. The number of proposal boxes are equal to the number of anchor boxes. Corresponding to these proposals boxes, features $[\textbf{f}^{\;p}_1, \textbf{f}^{\;p}_2, ..., \textbf{f}^{\;p}_{576}]$ are pooled from feature map $\textbf{F}$.

\textbf{Contrastive loss} is used for pre-training. The cosine distance $d_k$ is computed between each anchor feature $\textbf{f}^{\;anc}_k$ and $\textbf{f}^{\;p}_k$, the total number of features are $576$ for anchors and $576$ proposals. The pseudo label $y$ is of length $576$. $y_k$ is equal to $1$ where the $\textbf{f}^{\;anc}_k$ feature have potential object.
The scaling parameters for contrastive loss $\alpha$ and classification loss $\beta$ are set to $0.6, 0.4$ respectively. The network is trained for 100 epochs, convergence is observed around 45 epochs, based on that, early stopping is done at 50 epochs. The learning rate starts with $1e^{-5}$ with decay of 0.1 after every 10 epochs. The batch size is set at 24. The optimizer used is $Adam$ optimizer. The region proposal branch of the network learns to select features from regions where the objects are present. During training, we restrict the learning rate of linear projection layer of $f^{txt}_{ao}$ to a low value to stabilize the region proposal branch.

\textbf{Compositional Classifier} $CC(.)$: The ability to learn individual representation of O-A in visual domain is crucial for transferring knowledge from seen to unseen O-A associations. Existing SOTA works~\cite{naeem2021learning,mancini2021learning,mancini2021open,ruis2021independent,xu2021zero} use homogeneous features from whole image as without localizing the object, they ignore the discriminative visual features of object and its attributes. Our Composition Classifier network $CC(.)$ leverages the distinctive features extracted by $LFE(.)$ to predict the object and its corresponding attribute.
It is challenging to associate right attribute with the object by using homogeneous features, as there can be interference from prominent confounding elements.
$CC(.)$ takes as input, the top 10 pooled features $[\textbf{f}^{\;p}_1, \textbf{f}^{\;p}_2, ..., \textbf{f}^{\;p}_{10}]$ from pre-trained $LFE(.)$ sorted in descending order based on objectness score $\hat{o} = [\hat{o}_1, \hat{o}_2, .., \hat{o}_{10}]$. Each block in $CC(.)$ consists of two fully connected layer with ReLU activation. The initial learning rate for $CC(.)$ network is set to $1e^{-3}$ with a decay of 0.1 after every 7 epochs. We observed that fine tuning $LFE(.)$ with a lower learning rate of $1e^{-6}$ while training  $CC(.)$ performed better than freezing it. The batch size used is 32. All the experiments are done on a single nvidia V100 Tesla.

\subsection{Results \& Analysis}

We report Top-1 accuracy in seen and unseen classes and accuracy in detecting objects and attributes. LOCL achieve best accuracy in individual detection of objects and attributes as shown in Table~\ref{tab:main_result_obj_attr}. This is interesting as our simple $CC(.)$ do not a have dedicated object detector similar to SymNet~\cite{li2020symmetry}.
In MIT-States~\cite{isola2015discovering} LOCL outperforms SOTA method by $12\%$ on object detection accuracy and $19\%$ attribute detection. 
In UT-Zappos~\cite{yu2017semantic} LOCL's performance is slightly better than SOTA methods since each image has one dominant object with clear white background. Moreover the performance improvement is significant when it comes to more challenging and realistic dataset CGQA~\cite{naeem2021learning}. 

\begin{table*}[h]
\centering
\begin{adjustbox}{width=0.8\textwidth}
\begin{tabular}{l|cc|cc|cc}
\multicolumn{1}{c|}{\multirow{2}{*}{Methods}}        & \multicolumn{2}{c|}{MIT-States~\cite{isola2015discovering}} & \multicolumn{2}{c|}{UT-Zappos~\cite{yu2017semantic}} & \multicolumn{2}{c}{CGQA~\cite{naeem2021learning}} \\
\multicolumn{1}{c|}{}                                & Object                       & Attribute                    & Object                    & Attribute                & Object                  & Attribute               \\ \hline
\tablelightgray Attop~\cite{nagarajan2018attributes} & 21.1                         & 23.6                         & 38.9                      & 69.6                     & 8.3                     & 12.5                    \\
LabelEmbed~\cite{misra2017red}                       & 23.5                         & 26.3                         & 41.2                      & 69.2                     & 7.4                     & 15.6                    \\
\tablelightgray TMN~\cite{purushwalkam2019task}      & 23.3                         & 26.5                         & 40.8                      & 69.9                     & 9.7                     & 20.5                    \\
SymNet~\cite{li2020symmetry}                         & 26.3                         & 28.3                         & 40.5                      & 71.2                     & 14.5                    & 20.2                    \\
\tablelightgray CompCos~\cite{mancini2021open}       & 27.9                         & 31.8                         & 44.7                      & 73.5                     & -                       & -                       \\
CGE~\cite{naeem2021learning}                         & 30.1                         & 34.7                         & 48.7                      & 76.2                     & 15.2                    & 30.4                    \\
\tablelightgray LOCL (Ours)                          & \textbf{42.7}                & \textbf{53.4}                & \textbf{49.4}             & \textbf{79.3}            & \textbf{28.7}           & \textbf{35.1}           \\ \hline
\end{tabular}

\end{adjustbox}
\vspace{0.1cm}
 \caption{Performance comparisons on detecting individual objects and attributes. LOCL outperforms all compared methods with a significant margin.}
\label{tab:main_result_obj_attr}
\vspace{-0.4cm}
\end{table*}

\subsection{Datasets}
The splits used on all the datasets are as follows. MIT-States~\cite{isola2015discovering} has a total of 53,000 images with 245 objects and 115 attributes. The splits for MIT-States dataset have 1262 object-attribute pairs (34,000 images) for the training set, 600 object-attribute pairs (10,000 images) as the validation set and 800 pairs (12,000 images) as test set.  All the images in MIT-states dataset are of natural objects collected using an older search engine with limited human annotation causing a significant label noise~\cite{Atzmon_casual}.
UT-Zappos~\cite{yu2017semantic} has 29,000 images of shoes catalogue. The splits used are of 83 object-attribute pairs (23,000 images) for the training set, 30 object-attribute pairs (3,000 images) for the validation set and 36 pairs (3,000 images) for test set.  The images in UT-Zappos~\cite{yu2017semantic} dataset are not really entirely a compositional dataset as the attributes like \textit{Faux Leather vs Leather} are material differences but not specifically any visual difference~\cite{mancini2021learning}. Also, the simplicity of the images (one object with white background) makes it unsuitable to work in natural surroundings where the object of interest has interference confounding elements in the scene.  These splits are selected following previous works~\cite{purushwalkam2019task, mancini2021learning}.
The third dataset used is Compositional-GQA (CGQA) dataset~\cite{hudson2019gqa, naeem2021learning}. It has 453 attributes and 870 objects. The splits for CGQA have 5592 object-attribute pairs (26,000 images) for training set, 2292 pairs (7,000 images) for validation set and 1811 pairs (5,000 images) for testing set. These splits are as proposed by~\cite{mancini2021learning}. The CGQA dataset have images curated from visual genome dataset~\cite{krishna2017visual} which comprises of images from natural and realistic settings. Most of the images in CGQA have an object of interest with confounding elements in the background, that makes it an extremely challenging dataset to evaluate CGQA models. 

\subsection{Ablation Study}
In this section, we discuss about the additional design choices for training of the Localized Feature Extractor $LFE(.)$ and composition classifier $CC(.)$.

\subsubsection{Number of Proposals:}
Table~\ref{tab:abl:num_prop} shows the selection criterion of number of proposal selected from pre-trained $LFE(.)$ the goes as input to $CC(.)$. With $r<10$, the proposals features miss regions of the object, which leads to poor performance. While when $r>10$, more background features are picked that suppress the prominent object and lead to drop in prediction quality.

\begin{table}[h]
\centering
\begin{tabular}{l|lllll}
\# of proposals & \multicolumn{3}{c}{MIT-States}               \\ 
               & Seen          & Unseen         & AUC          \\ \hline
\tablelightgray
5              & 32.1         & 33.6          & 7.2          \\
10             & \textbf{35.3} & \textbf{36.0} & \textbf{7.7} \\
\tablelightgray
15             & 35.3         & 35.9          & 6.9          \\
20             & 27.6         & 28.4          & 6.5          \\ \hline
\end{tabular}
\vspace{0.1cm}
\caption{Performance of LOCL as we select different number of top \textit{r} proposals from pre-trained LFE. Best performance is observed with \textit{r=10}. With $r>10$, more background features are picked that suppress the prominent objects.}
\label{tab:abl:num_prop}
\end{table}

\subsubsection{Object$\backslash$Attribute Refinement:}
Table~\ref{tab:abl:refinement} shows refinement operations done on visual features $\textbf{f}^{\;all}_p$ as shown in Eq.6 in the main paper. The multiplication operations generates more selective information and suppresses the redundant information as compared to concatenation and addition operation~\cite{ulutan2020vsgnet,iftekhar2021gtnet}.

\begin{table}[h]
\centering
\begin{tabular}{l|lllll}

Method         & \multicolumn{3}{c}{MIT-States}~\cite{isola2015discovering}             \\ 
               & Seen          & Unseen        & AUC          \\ \hline
\tablelightgray
Addition       & 28.5          & 29.6          & 6.6          \\
Multiplication & \textbf{35.3} & \textbf{36.0} & \textbf{7.7} \\
\tablelightgray
Concatenation  & 32.7             & 33.1             & 7.2            \\ \hline
\end{tabular}
 \caption{Performance of compositional classifier with different refinement operations. }
\label{tab:abl:refinement}
\end{table}

\subsubsection{Pre-training $LEF(.)$ with object embeddings:} As discussed in section 3.1 of the main paper, we use OA pair name $<Blue,Bird>$ as input during the pre-training of $LEF(.)$. We also test with using only the object names $<Bird>$ as the input. We observe $3\%$ drop in accuracy as compared to OA pair names in the unseen category as shown in Table~\ref{tab:obj}. This is expected as the text embeddings generated by the text encoder are more meaningful and have closer representation with the visual features when we provide a complete description of the object in the image i.e. OA pair name.

\begin{table}[h]
\centering
\begin{tabular}{l|lllll}
\multirow{2}{*}{Names used}      & \multicolumn{5}{c}{MIT-States~\cite{isola2015discovering}}                                                                            \\
                                 & \multicolumn{1}{l|}{Seen} & \multicolumn{1}{l|}{Unseen} & \multicolumn{1}{l|}{Obj}  & \multicolumn{1}{l|}{Attr} & \textit{AUC} \\ \hline
\tablelightgray\textit{Obj-Attr} & \multicolumn{1}{l|}{\textbf{35.3}} & \multicolumn{1}{l|}{\textbf{36.0}}   & \multicolumn{1}{l|}{\textbf{42.7}} & \multicolumn{1}{l|}{\textbf{53.4}} & \textbf{7.7} \\
\textit{Obj}                     & \multicolumn{1}{l|}{32.5} & \multicolumn{1}{l|}{32.8}   & \multicolumn{1}{l|}{37.4} & \multicolumn{1}{l|}{41.9} & 7.1         
\end{tabular}
\vspace{0.1cm}
\caption{Performance of the network in MIT-States~\cite{isola2015discovering} with different names used as input to the text encoder while pre-training $LFE(.)$. \textbf{Bold} numbers are the best performance settings. The network performs well with $Obj-Attr$ names as input compared to just $Obj$ names.} 
\label{tab:obj}
\vspace{-0.4cm}
\end{table}

\subsubsection{Number of Pseudo Labels:} For creating pseudo labels $y$  during pre-training, as mentioned in section 3.1 equation 4, we assign value 1 to top 20 indexes and rest are assigned 0. The equation is:
\begin{equation}
    \phi = [\phi_1, \phi_2, .., \phi_k, .., \phi_n]
    \label{eq:pseudo_label2}
\end{equation}

\begin{equation}
  y =
    \begin{cases}
      1 & argsort(\phi)[0:l]\\
      0 & for\; all\; other\;indexes\\
    \end{cases}
    \label{eq:pseudo_label}
\end{equation}
where $y = [y_1, y_2, .., y_k, .., y_n]$, $20$ anchors are selected based on cosine similarity score $\phi$ (equation 2 in main paper). They are assigned with label 1 in $y$ and rest are assigned 0 as shown above with Eq.~\ref{eq:pseudo_label}. Here each $y_k$ represents the presence/absence of object of interest regions in the input image. 
We experiment with different values for number of potential objects. As shown in Table~\ref{tab:psd}, the overall performance of the model drops if we pick a number greater than or less than $20$. This is because for smaller value, the $LFE(.)$ is penalized for detecting even the right regions of interests and for larger value than 20, we are learning information from confounding elements from the background where the object may/may not be present.

\begin{table}[h]
\centering
\begin{tabular}{c|ccccc}

\multirow{2}{*}{\begin{tabular}[c]{@{}c@{}}\#Pseudo \\ Labels\end{tabular}} & \multicolumn{5}{c}{MIT-States~\cite{isola2015discovering}}                                                                                                                                    \\ 
                                                                            & \multicolumn{1}{c|}{Seen} & \multicolumn{1}{c|}{Unseen} & \multicolumn{1}{c|}{AUC} & \multicolumn{1}{c|}{Obj}  & Attr \\ \hline
\tablelightgray
\textit{10}                                                                 & \multicolumn{1}{c|}{31.5}          & \multicolumn{1}{c|}{27.9}            & \multicolumn{1}{c|}{5.2}          & \multicolumn{1}{c|}{27.9}          & 31.5          \\ 
\textit{15}                                                                 & \multicolumn{1}{c|}{33.8}          & \multicolumn{1}{c|}{34.1}            & \multicolumn{1}{c|}{6.5}          & \multicolumn{1}{c|}{28.4}          & 30.8          \\ 
\tablelightgray
\textit{20}                                                        & \multicolumn{1}{c|}{\textbf{35.3}} & \multicolumn{1}{c|}{\textbf{36.0}}   & \multicolumn{1}{c|}{\textbf{7.7}} & \multicolumn{1}{c|}{\textbf{42.7}} & \textbf{53.4} \\ 
\textit{25}                                                                 & \multicolumn{1}{c|}{29.1}          & \multicolumn{1}{c|}{29.6}            & \multicolumn{1}{c|}{6.1}          & \multicolumn{1}{c|}{31.5}          & 34.6          \\ 
\end{tabular}
\vspace{0.1cm}
\caption{Performance of the network in MIT-States~\cite{isola2015discovering} with different number of region of interest while pre-training $LFE(.)$. \textbf{Bold} numbers are the best performance settings.  Here \# is "Number of".} 
\label{tab:psd}
\vspace{-0.4cm}
\end{table}

\subsubsection{Margin distance for contrastive loss:} For pre-training $LEF(.)$ with contrastive loss, we use a margin distance of $1$ as shown in equation 5 in the main paper. We experimented with different distances for the margin for MIT-states~\cite{isola2015discovering} dataset. We achieved best performance at a margin of 1. The experimental evaluation with different margin distance is shown in Table~\ref{tab:margin}. Our observations of is that with bigger margin, the network start clustering features from those regions also, which have object of interest along with a significant section of background regions. This leads to drop in attribute detection accuracy.

\begin{table}[h]
\centering
\begin{tabular}{l|lllll}

\multirow{2}{*}{\textit{Margin}} & \multicolumn{5}{c}{MIT-States~\cite{isola2015discovering}}                                                                                                                                  \\ 
                                 & \multicolumn{1}{l|}{Seen}          & \multicolumn{1}{l|}{Unseen}        & \multicolumn{1}{l|}{AUC}          & \multicolumn{1}{l|}{Object}        & Attribute     \\ \hline
\tablelightgray
0.5                              & \multicolumn{1}{l|}{29.6}          & \multicolumn{1}{l|}{30.4}          & \multicolumn{1}{l|}{5.2}          & \multicolumn{1}{l|}{30.2}          & 47.3          \\ 
1.0                              & \multicolumn{1}{l|}{\textbf{35.3}} & \multicolumn{1}{l|}{\textbf{36.0}} & \multicolumn{1}{l|}{\textbf{7.7}} & \multicolumn{1}{l|}{\textbf{42.7}} & \textbf{53.4} \\ 
\tablelightgray
3                                & \multicolumn{1}{l|}{34.1}          & \multicolumn{1}{l|}{33.9}          & \multicolumn{1}{l|}{6.5}          & \multicolumn{1}{l|}{41.1}          & 46.8          \\ 
7                                & \multicolumn{1}{l|}{25.3}          & \multicolumn{1}{l|}{26.5}          & \multicolumn{1}{l|}{4.8}          & \multicolumn{1}{l|}{37.3}          & 38.9          \\ 
\end{tabular}
\vspace{0.1cm}
\caption{Performance of pre-training the $LFE(.)$ using different margin distance for contrastive learning. We achieve best performance when margin is $1$. For higher margin, $LFE(.)$ cluster features of object of interest which have significant region of background/confounding regions also. Leaning to poor performance.}
\label{tab:margin}
\vspace{-0.4cm}
\end{table}

\subsubsection{Scaling parameters of loss function:} While pre-training, we combine contrastive loss and binary cross entropy loss using scaling parameters $\alpha$ and $\beta$. The equation is:
\begin{equation}
    \mathcal{L}_{total} = \alpha * \mathcal{L}_{CON} + \beta * \mathcal{L}_{BCE}(o, \phi),
    \label{eq:loss_cl}
\end{equation}
where, $\mathcal{L}_{CON}$ is the contrastive loss and $\mathcal{L}_{BCE}$ is the binary cross entropy loss. We test for different values $\alpha$ and $\beta$ as shown in Table~\ref{tab:alpha_beta}. It appears giving a bit more weight to the contrastive loss helps $LFE(.)$ to extract better localized features. 

\begin{figure*}[]
    \includegraphics[width=\linewidth, height=7in]{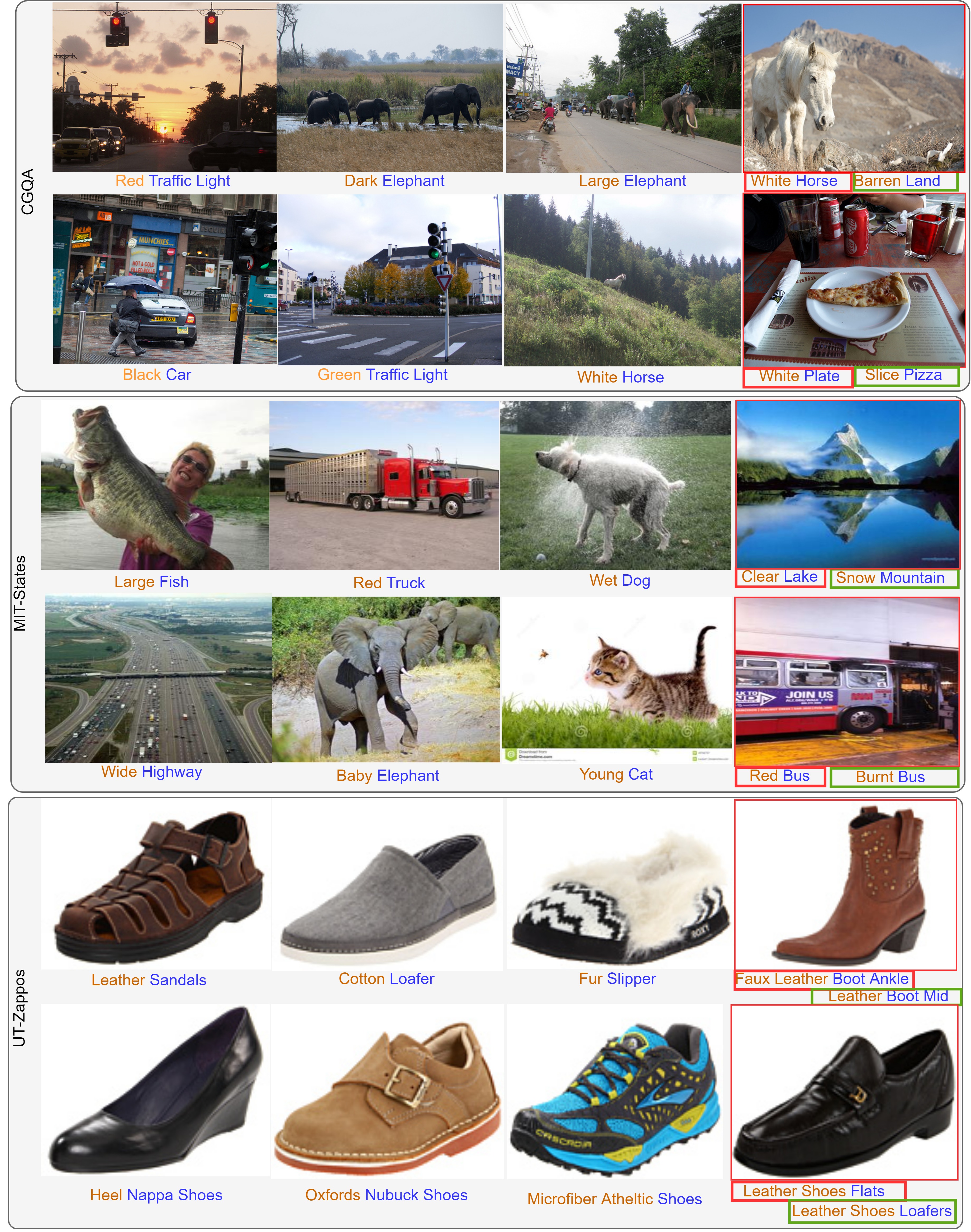}
    \caption{Qualitative results of LOCL. Left three columns show correct predictions from our network. Rightmost column shows missed predictions, here, ground truth labels are marked with green box and our predictions are marked in red box. The datasets contain only one OA pair and our predictions though visually correct, do not match with the ground-truth OA in these cases.}
    \label{fig:Qual-Results-sup}
\end{figure*}

\begin{table}[h]
\centering
\begin{tabular}{cc|cccll}

\multicolumn{2}{c|}{Parameters}                                   & \multicolumn{5}{c}{MIT-States~\cite{misra2017red}}                                                                                                                                    \\ 
\multicolumn{1}{c|}{$\alpha$}                     & $\beta$                     & \multicolumn{1}{c|}{\textit{Seen}} & \multicolumn{1}{c|}{\textit{Unseen}} & \multicolumn{1}{c|}{\textit{AUC}} & \multicolumn{1}{l|}{Obj}           & Attr          \\ \hline
\tablelightgray
\multicolumn{1}{c|}{\textit{0.3}}          & \textit{0.7}          & \multicolumn{1}{c|}{30.6}          & \multicolumn{1}{c|}{32.5}            & \multicolumn{1}{c|}{6.9}          & \multicolumn{1}{l|}{30.9}          & 33.1          \\ 
\multicolumn{1}{c|}{\textit{0.4}}          & \textit{0.6}          & \multicolumn{1}{c|}{35.1}          & \multicolumn{1}{c|}{35.4}            & \multicolumn{1}{c|}{7.5}          & \multicolumn{1}{l|}{36.4}          & 39.9          \\ 
\tablelightgray
\multicolumn{1}{c|}{\textit{0.5}}          & \textit{0.5}          & \multicolumn{1}{c|}{34.9}          & \multicolumn{1}{c|}{35.8}            & \multicolumn{1}{c|}{7.7}          & \multicolumn{1}{l|}{40.8}          & 49.3          \\ 
\multicolumn{1}{c|}{\textit{\textbf{0.6}}} & \textit{\textbf{0.4}} & \multicolumn{1}{c|}{\textbf{35.3}} & \multicolumn{1}{c|}{\textbf{36.0}}   & \multicolumn{1}{c|}{\textbf{7.7}} & \multicolumn{1}{l|}{\textbf{42.7}} & \textbf{53.4} \\ 
\tablelightgray
\multicolumn{1}{c|}{\textit{0.7}}          & \textit{0.3}          & \multicolumn{1}{c|}{32.7}          & \multicolumn{1}{c|}{33.7}            & \multicolumn{1}{c|}{7.1}          & \multicolumn{1}{l|}{33.0}          & 35.2          \\ 
\end{tabular}
\vspace{0.1cm}
\caption{Performance of the network with different scaling parameters of the loss function during pre-training. \textbf{Bold} numbers are the best performance settings.} 
\label{tab:alpha_beta}
\vspace{-0.4cm}
\end{table}


\subsection{Qualitative Results}
We add more qualitative results for unseen novel composition with top-1 prediction in Figure~\ref{fig:Qual-Results-sup}. The examples are presented from datasets : CGQA~\cite{naeem2021learning}, MIT-States~\cite{isola2015discovering}, and UT-Zapos~\cite{yu2017semantic}. The order of the datasets is in decreasing order of the clutter in the images. As can be seen that in the CGQA dataset, the images contains object of interest with lot of confounding elements creating background clutter. MIT-States~\cite{isola2015discovering} is also of natural images. However, most of the images have a dominant object. On the other hand, in UT-Zappos~\cite{yu2017semantic} all the images contain a single object with clear white background. This shows the complexity and the challenges of CGQA dataset compared to the existing ones. 

The first three columns represent the examples where our model is making the right predictions. The last column in each dataset shows examples where our model makes the visually correct prediction. However, it does not match with the ground truth label of object and attribute. Our model is selecting object of interest, and it is creating the right attribute-object associations. For example in case of fourth row on the rightmost column, our prediction of the object is right but the image contains multiple attributes, while the ground truth contains only one OA pair. This put an artificial limitation on the evaluation metric even when the predictions are perceiveably correct. 

\paragraph{BMP-Net~\cite{xu2021relation}}achieves state-of-the-art (SOTA) performance in seen classes of MIT-States~\cite{misra2017red} and UT-Zappos~\cite{yu2017semantic}. However, their sub-optimal performance in unseen classes indicates a bias towards seen classes. To further investigate this bias, we evaluate BMPNet on the challenging CGQA dataset~\cite{naeem2021learning}. We utilize the official repository provided by the authors for this evaluation and report performance in the same matrices used for other datasets. In Table~\ref{tab:bmpnet}, we can observe LOCL outperforms BMPNet in all category. Especially in unseen classes, LOCL achieves more than double accuracy than BMPNet. This poor performance indicates a seen class bias of BMPNet. This bias is mainly due to creating the graph network with a large number of seen and plausible OA pairs. More discussion on this phenomenon is available in section 4.4 in the main paper. Moreover, LOCL is very efficient and utilizes only $\sim5$GB memory for training in the large-scale dataset CGQA. Current graph-based SOTA networks CGE~\cite{naeem2021learning} ($\sim10$GB), BMPNet~\cite{xu2021relation} ($\sim40$GB) utilize much higher GPU memory for the same batch size in CGQA dataset. Therefore, LOCL is suitable for training on large scale challenging CZSL datasets.

\begin{table}[h]
\centering
\begin{tabular}{l|ccc}
\multicolumn{1}{c|}{\multirow{2}{*}{Methods}} & \multicolumn{3}{c}{CGQA~\cite{naeem2021learning}}                      \\
\multicolumn{1}{c|}{}                         & \multicolumn{1}{c|}{Seen} & \multicolumn{1}{c|}{Unseen} & AUC          \\ \hline
\tablelightgray BMP-Net~\cite{xu2021relation} & \multicolumn{1}{c|}{29.1} & \multicolumn{1}{c|}{11.7}   & 2.7          \\
LOCL (Ours)                                   & \multicolumn{1}{c|}{\textbf{29.6}} & \multicolumn{1}{c|}{\textbf{26.4}}   & \textbf{4.2}
\end{tabular}
\vspace{0.1cm} 
\caption{Performance comparison on CGQA~\cite{naeem2021learning} dataset. LOCL significantly outperform BMP-Net~\cite{xu2021relation} in a challenging (significant background clutter) dataset. The performance of LOCL shows the effectiveness of \textbf{LEF} in unseen OA associations.}
\label{tab:bmpnet}
\vspace{-0.4cm}
\end{table}